\documentclass[11pt]{article}

\usepackage[preprint]{acl}
\usepackage{amsmath}
\usepackage{booktabs}    
\usepackage{multirow}    
\usepackage{multicol}    
\usepackage{graphicx}    
\usepackage{colortbl}   
\usepackage{xcolor}     
\usepackage{times}
\usepackage{siunitx}
\usepackage{latexsym}
\usepackage{caption}
\usepackage{float}
\usepackage[T1]{fontenc}

\usepackage[utf8]{inputenc}

\usepackage{microtype}

\usepackage{inconsolata}

\usepackage{graphicx}
\usepackage{fontawesome5}
\newcommand*{\imgintext}[1]{%
  \raisebox{-0.2ex}{%
    \includegraphics[
      height=2.3ex, 
      keepaspectratio
    ]{#1}%
  }%
}

\newcommand*{\imgintitle}[1]{%
  \raisebox{-2.5ex}[0pt][0pt]{
    \includegraphics[
      height=5ex, 
      keepaspectratio
    ]{#1}%
  }%
}

\definecolor{bestcolor}{HTML}{E6D9F0}   
\definecolor{secondcolor}{HTML}{FFE5CC} 
\definecolor{thirdcolor}{HTML}{D9D9D9}  

\definecolor{myred}{HTML}{DC143C}     
\definecolor{mygreen}{HTML}{2E8B57}   
\definecolor{myyellow}{HTML}{DAA520}  

\newcommand{\best}[1]{\cellcolor{bestcolor}\textbf{#1}}
\newcommand{\second}[1]{\cellcolor{secondcolor}\textbf{#1}}
\newcommand{\third}[1]{\cellcolor{thirdcolor}\textbf{#1}}

\usepackage{pifont} 
\newcommand{\yes}{\textcolor{mygreen}{\ding{52}}}
\newcommand{\no}{\textcolor{myred}{\ding{55}}}
\newcommand{\half}{\textcolor{myyellow}{\ding{52}\rotatebox[origin=c]{-9.2}{\kern-0.7em\ding{55}}}} 

\newcommand{\TCMBENCH}{TCM-Eval} 
\title{\imgintitle{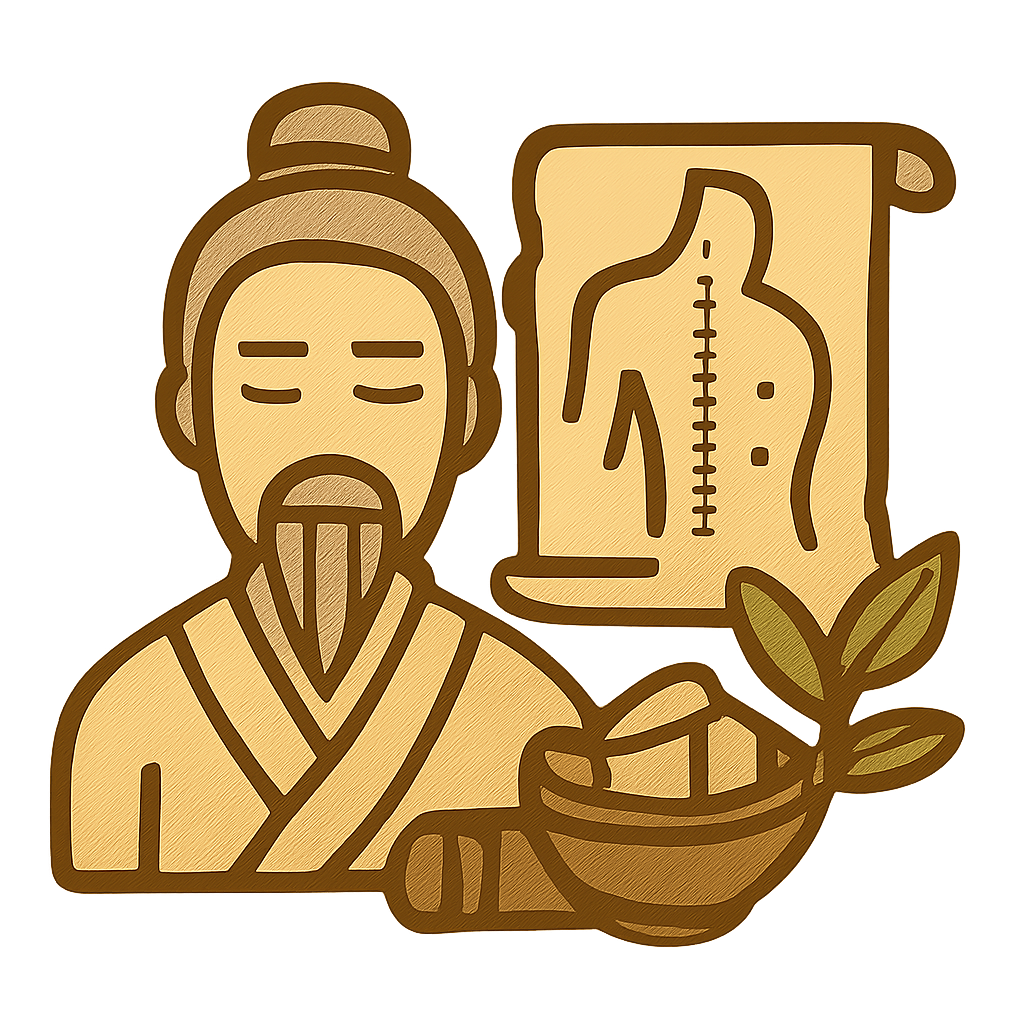} TCM-Eval: An Expert-Level Dynamic and Extensible Benchmark for Traditional Chinese Medicine}

\author{
  \textbf{Zihao Cheng}$^{1^*}$ \thanks{$^*$Equal contribution and the order is determined alphabetically by last name.},
  \textbf{Yuheng Lu}$^{1^*}$ ,
  \textbf{Huaiqian Ye}$^{1}$,
  \textbf{Zeming Liu}$^{1\dagger}$ \thanks{$^\dagger$Corresponding author: Zeming Liu.},
  \textbf{Minqi Wang}$^{2}$,\\
  \textbf{Jingjing Liu}$^{1}$,
  \textbf{Zihan Li}$^{1}$,
  \textbf{Wei Fan}$^{2}$,
  \textbf{Yuanfang Guo}$^{1}$,
  \textbf{Ruiji Fu}$^{3\ddagger}$
  \thanks{$^\ddagger$Project leader: Ruiji Fu.},
  \textbf{Shifeng She}$^{2,4}$,\\
  \textbf{Gang Wang}$^{2}$,
  \textbf{Yunhong Wang}$^{1}$,\\
  $^{1}$School of Computer Science and Engineering, Beihang University\\
  $^{2}$Beijing Zhimingtang Technology Co., Ltd. 
  $^{3}$Beijing Zhiyan AI Technology Co., Ltd.\\
  $^{4}$Guangzhou University of Chinese Medicine\\
  \faGlobe~\url{https://tcmeval.bamaidical.com}
  \\[0.5em]
}

\makeatletter
\def\thanks#1{\protected@xdef\@thanks{\@thanks
        \protect\footnotetext{#1}}}
\makeatother

\begin{document}
\maketitle

\begin{abstract}

Large Language Models (LLMs) have demonstrated remarkable capabilities in modern medicine, yet their application in Traditional Chinese Medicine (TCM) remains severely limited by the absence of standardized benchmarks and the scarcity of high-quality training data. To address these challenges, we introduce \textbf{TCM-Eval}, the first dynamic and extensible benchmark for TCM, meticulously curated from national medical licensing examinations and validated by TCM experts. Furthermore, we construct a large-scale training corpus and propose \textbf{S}elf-\textbf{I}terative \textbf{C}hain-\textbf{o}f-\textbf{T}hought \textbf{E}nhancement (\textbf{SI-CoTE}) to autonomously enrich question-answer pairs with validated reasoning chains through rejection sampling, establishing a virtuous cycle of data and model co-evolution. Using this enriched training data, we develop \textbf{Z}hi\textbf{M}ing\textbf{T}ang-\textbf{M1} \textbf{(ZMT-M1)}, a state-of-the-art LLM specifically designed for TCM, which significantly exceeds the passing threshold for human practitioners. To encourage future research and development, we release a public leaderboard, fostering community engagement and continuous improvement.

\end{abstract}
\section{Introduction}

\begin{figure}
    \centering
    \includegraphics[width=0.45\textwidth]{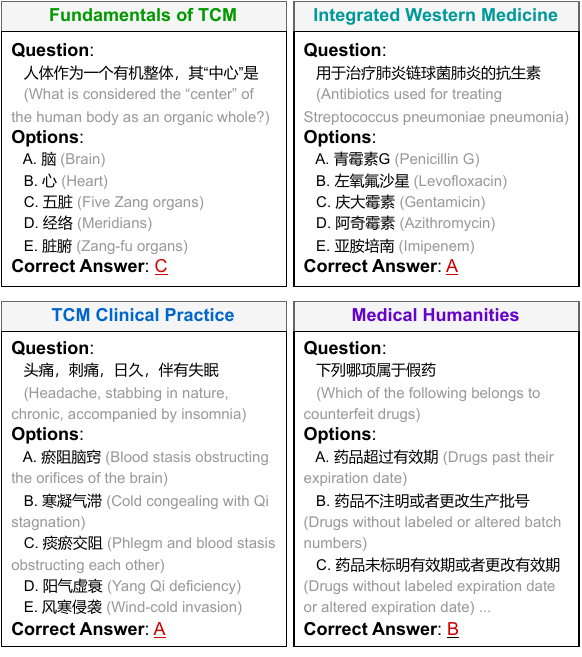}
    \caption{Examples of \textbf{TCM-Eval}, illustrating four aspects: Fundamentals of TCM, TCM Clinical Practice, Integrated Western Medicine, Medical Humanities.}
    \label{fig:data_example}
\end{figure}

The rapid advancement of Large Language Models (LLMs) has catalyzed a paradigm shift across numerous domains \cite{zhao2023survey, gao2025survey, survey_data_synthesis}, with healthcare emerging as a particularly promising frontier \cite{thirunavukarasu2023large, zhou2023survey}. These models have demonstrated remarkable capabilities in tasks such as clinical decision support \cite{li2025questionsclinicalrecommendationslarge, garza2025retrievalaugmentedframeworkllmbasedclinical}, medical text summarization \cite{van2024adapted, aali2025dataset}, and patient-facing conversational AI \cite{mukherjee2024polaris}, significantly accelerating innovation in the medical field.

However, despite these advancements, the focus of current medical LLMs and open-source datasets has been overwhelmingly skewed towards modern medicine \cite{liu2024medbenchcomprehensivestandardizedreliable, wu2024medjourney, chen2024clinicalbench, zhang2025llmeval}. This has created a significant gap in the domain of Traditional Chinese Medicine (TCM). The development of powerful LLMs for TCM is severely hampered by two fundamental challenges: the absence of a unified, authoritative platform for model evaluation \cite{chen2025shizhengptmultimodalllmstraditional}, and the scarcity of high-quality, large-scale training data \cite{qiu2024buildingmultilinguallanguagemodel, zhang2024qibolargelanguagemodel}. This lack of foundational resources restricts progress and prevents a systematic understanding of current models' capabilities within the TCM domain.

To address these challenges, we introduce \textbf{TCM-Eval}, the first dynamic and extensible benchmark specifically designed for evaluating LLMs in the domain of TCM. TCM-Eval consists of 6,099 high-quality questions, carefully selected from a decade's worth of the National Medical Licensing Examination for TCM Practitioners. Each question has been validated by experts in TCM to ensure its accuracy and relevance. 
As shown in Figure \ref{fig:data_example}, TCM-Eval comprehensively tests the models' capabilities across four key dimensions: \textit{(1) Fundamentals of TCM, (2) TCM Clinical Practice, (3) Integrated Western Medicine, and (4) Medical Humanities}. All questions are formatted as multiple-choice, ensuring the uniqueness of answers and the verifiability of responses.
To maintain the benchmark's integrity and prevent data leakage, TCM-Eval is dynamic, with continuous updates to incorporate new questions.

To address the challenge of scarce high-quality training data, we have compiled a comprehensive corpus from 18 authoritative TCM textbooks and thousands of mock examinations, which we used to create a domain-specific dataset consisting of over 384,807 question-answer pairs. To leverage this data and enhance reasoning abilities, we introduce \textbf{S}elf-\textbf{I}terative \textbf{C}hain-\textbf{o}f-\textbf{T}hought \textbf{E}nhancement \textbf{(SI-CoTE)}, designed to autonomously augment simple QA pairs with high-quality, step-by-step reasoning processes. The SI-CoTE process operates iteratively: the model generates potential reasoning chains for a subset of the data, and through Rejection Sampling \cite{liu2023statistical}, we retain only those chains that lead to the correct answer. This newly generated high-quality CoT data is then used to fine-tune the model, yielding a more capable version. This enhanced model, in turn, processes the next subset of data, creating a virtuous cycle where both the training data quality and the model's reasoning abilities evolve synergistically. Using this efficient self-improvement strategy, we developed \textbf{Z}hi\textbf{M}ing\textbf{T}ang-\textbf{M1 }\textbf{(ZMT-M1)} by fine-tuning the powerful Deepseek-R1 \cite{deepseekai2025deepseekr1incentivizingreasoningcapability} foundation model.

We conducted extensive experiments on TCM-Eval using both open-source and closed-source models. ZMT-M1 achieved an impressive average score of 96.32, significantly outperforming all 34 competing models. It also far exceeds the 60-point passing score required for human practitioners in the official examination. This demonstrates that ZMT-M1 has attained expert-level knowledge and reasoning abilities in TCM, matching or even surpassing the capabilities of qualified human professionals. Additionally, we observed a significant data leakage issue in current models, where performance on newly added questions was notably lower compared to older questions. In contrast, ZMT-M1 showed no such decline, maintaining consistent performance across both old and new questions.

In summary, our key contributions are threefold:
\begin{itemize}
    \item We introduce \textbf{TCM-Eval}, the first dynamic, extensible, and expert-validated benchmark specifically designed for TCM. 

    \item We build a high-quality QA dataset from mock exams and authoritative TCM textbooks. Using the \textbf{SI-CoTE} method, we expanded it into QA pairs with reasoning chains. This enriched dataset is used to train \textbf{ZMT-M1}, establishing a new SOTA in the TCM domain.

    \item We establish the first comprehensive evaluation platform for TCM LLMs, offering a standardized testbed and a public leaderboard to drive research in the field.
\end{itemize}
\section{Related Work}

\begin{table*}[!htbp]
\centering
\resizebox{\textwidth}{!}{%
\begin{tabular}{@{}lccccccr@{}}
\toprule[0.08em]
\multirow{2}{*}{\textbf{Datasets}} & \multirow{2}{*}{\textbf{Dynamic}} & \textbf{TCMLE-} & \multirow{2}{*}{\textbf{Textbook}} & \textbf{Real} & \textbf{Mock} & \textbf{Expert-} & \multirow{2}{*}{\textbf{Scale}} \\ 
 &  & \textbf{Specific} &  & \textbf{Exam} & \textbf{Exam} & \textbf{Reviewed} &    \\ \midrule
TCM-Bench \cite{yueTCMBenchComprehensiveBenchmark2024} & \no & \yes & \no & \yes & \no & \half &  5,473 \\
TCM-3CEval \cite{huangTCM3CEvalTriaxialBenchmark2025} & \no & \no & \yes & \no & \no & \yes &  450 \\
MTCMB \cite{kongMTCMBMultiTaskBenchmark2025} & \no & \no & \no & \yes & \no & \yes &  7,100 \\
TCMEval-SDT \cite{wangTCMEvalSDTBenchmarkDataset2025} & \no & \no & \yes & \no & \no & \yes &  300 \\
TCM-Ladder \cite{xieTCMLadderBenchmarkMultimodal2025} & \no & \no & \yes & \yes & \no & \yes  & 52K \\
OphthBench \cite{zhouOphthBenchComprehensiveBenchmark2025} & \no & \half & \yes & \yes & \no & \yes  & 591 \\
\textbf{TCM-Eval(Ours)} & \yes & \yes & \yes & \yes & \yes & \yes & 6,099 \\ \bottomrule
\end{tabular}
}
\caption{\textbf{Overview of Datasets in the TCM Domain.} TCMLE-Specific refers to datasets specifically designed for the TCM Licensing Examination.}
\label{tab:tcm_datasets}
\end{table*}

\subsection{Large Language Models for Traditional Chinese Medicine}
In recent years, a series of Large Language Models for TCM have emerged, which can be categorized into three types: (1) Large language models based on external TCM knowledge bases; (2) Large language models with basic conversational capabilities in TCM; (3) Large language models with expert-level reasoning capabilities in TCM.

One type of research integrates knowledge from external Traditional Chinese Medicine (TCM) knowledge bases into open-source large language models to assist the models in generating responses. 
TCM-KLLaMA \cite{ZHUANG2025109887} extracts knowledge from knowledge graphs and injects it into the input text of the model; BenTsao \cite{wang2023huatuo}, OpenTCM \cite{he2025opentcmgraphragempoweredllmbasedtraditional} utilizes knowledge graph based retrieval enhanced generation technology to explicitly utilize knowledge from the knowledge graph bases during inference. 

However, they remain confined to guidance via simple prompt engineering or reliance on retrieval from external knowledge bases to assist decision-making, and these approaches have not truly expanded the inherent knowledge boundaries of TCM LLMs.

One type of research applies simple question-answer pairs for Supervised Fine-Tuning (SFT) \cite{Ding2023ParameterefficientFO} on open-source large language models, enabling the models to acquire TCM domain knowledge.
TCMLLM \cite{haoyu2024tcmllm}, TCM-FTP \cite{zhou2024tcmftpfinetuninglargelanguage}, ShenNong TCM-LLM \cite{zhu2023ChatMed}, BianCang \cite{Wei2024BianCang}, TCMChat \cite{DAI2024107530} are fine-tuned on instructions based on TCM knowledge and Chinese question answering; BianQue \cite{chen2023bianque} , Qibo \cite{JIA2025127672} and CMLM-ZhongJing \cite{Kang2025} build multi-turn dialogue datasets based on doctor-patient roles and inquiry scenarios to enhance their consultation capabilities.

However, these models only suitable for simple dialogue scenarios. Plagued by low-quality fine-tuning data, they are prone to hallucinations and thus cannot truly match the reasoning capabilities of human TCM practitioners. In contrast, through Rejection Sampling, our \textbf{SI-CoTE} automatically validate and retain the correct ground-truth answer, thereby augmenting simple QA pairs with high-quality, step-by-step reasoning processes.

The other type of research adopts chain-of-thought (CoT) \cite{wei2023chainofthoughtpromptingelicitsreasoning} data for SFT on open-source large language models, allowing the models to not only grasp the basic knowledge of the TCM domain but also possess the reasoning capabilities of TCM domain experts.  
Qibo \cite{JIA2025127672} and Lingdan \cite{hua2024lingdan} construct Chain-of-Thought for TCM consultation prompts, endowing the models with preliminary thinking logic and reasoning capabilities; JingFang \cite{yang2025jingfangexpertlevellargelanguage} improves the model's ability in comprehensive clinical consultation and precise syndrome differentiation through the design of Multi-Agent Collaborative Chain-of-Thought Mechanism (MACCTM) and Dual-Stage Recovery Scheme (DSRS). 

However, these methods only apply static CoT data to prompts for model reasoning, failing to form dynamic iteration of CoT data.In contrast, our ZMT-M1 autonomously enriches question-answer pairs with validated reasoning chains through rejection sampling, establishing a virtuous cycle of data and model co-evolution.

\subsection{Datasets for Traditional Chinese Medicine}
\begin{figure*}[!htbp]
    \centering
    \includegraphics[width=\textwidth]{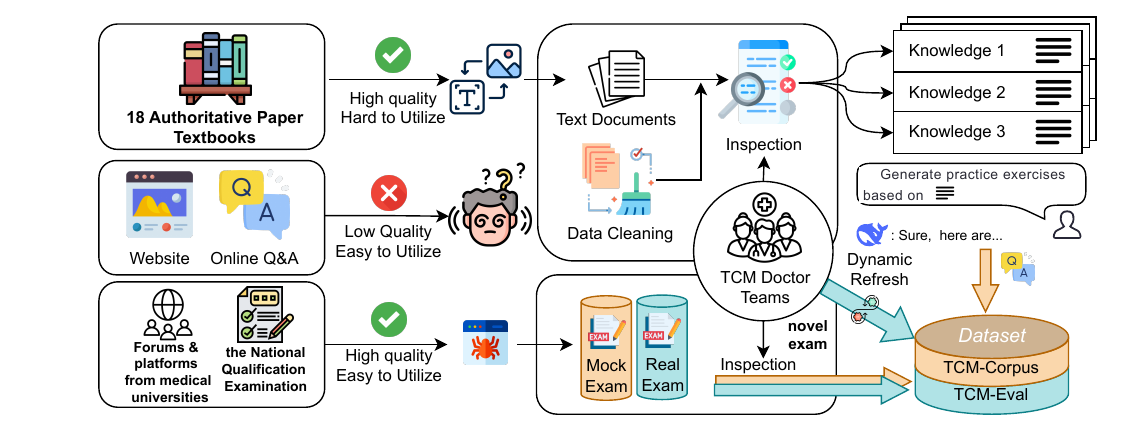}
    \caption{\textbf{Data Collection Pipeline.} Training Data is used for model training, and Test Data is continuously updated for model evaluation. Ordinary websites and online Q\&A  were excluded to ensure data quality.}
    \label{fig:data_collection}
\end{figure*}

In recent years, evaluation benchmarks specifically designed for the TCM domain have emerged. TCMBench \cite{yueTCMBenchComprehensiveBenchmark2024} targets practitioner exams across 16 knowledge areas, while TCM-3CEval \cite{huangTCM3CEvalTriaxialBenchmark2025} and MTCMB \cite{kongMTCMBMultiTaskBenchmark2025} broaden the scope to include literature understanding, diagnostic reasoning, and prescription recommendation. Benchmarks such as TCMEval-SDT \cite{wangTCMEvalSDTBenchmarkDataset2025} and TCM-Ladder \cite{xieTCMLadderBenchmarkMultimodal2025} emphasize structured and multimodal reasoning, with the latter incorporating hierarchical difficulty and text–image inputs. Comprehensive frameworks like TCMBench \cite{yueTCMBenchComprehensiveBenchmark2024} cover both theory and clinical decision-making, and cross-domain efforts such as OphthBench \cite{zhouOphthBenchComprehensiveBenchmark2025} demonstrate adaptability to specialized fields.
However, most existing benchmarks are static and narrow in coverage limiting their long-term applicability.
In contrast, the TCM-Eval dataset introduces a dynamic, expert-validated benchmark based on NMLE-TCM, with a systematically constructed training corpus derived from 18 authoritative textbooks and mock exams.

\section{Data Collection}
Our proposed dataset can be divided into two parts: TCM-Eval, a benchmark for Traditional Chinese Medicine (TCM) developed with expert oversight and inter-rater consistency to ensure accuracy(Section~\ref{sec:test_data_collection}); and TCM-Corpus, a training corpus constructed via an expert--LLM collaborative pipeline enhanced by our Self-Iterative Chain-of-Thought Enhancement (SI-CoTE) framework, which was used to train our model (Section~\ref{sec:train_data_collection}). We apply rigorous quality control protocols to both parts (Section~\ref{sec:data_quality_control}), and report their distributions, domain coverage, and key statistics (Section~\ref{sec:data_stat}).
\subsection{TCM-Eval} \label{sec:test_data_collection}
\begin{table*}[!htbp]
\small 
\centering
\begin{tabular}{@{}lcccccccccccc@{}} 
\toprule
\textbf{Unit} & \textbf{2003} & \textbf{2004} & \textbf{2007} & \textbf{2008} & \textbf{2009} & \textbf{2012} & \textbf{2013} & \textbf{2016} & \textbf{2022} & \textbf{2024} & \textbf{HC} & \textbf{Subtotal} \\
\midrule
Unit 1 & 135 & 131 & 150 & 149 & 100 & 150 & 150 & 150 & 150 & 142 & 150 & 1415 \\
Unit 2 & 135 & 142 & 150 & 150 & 100 & 150 & 150 & 150 & 150 & 105 & 150 & 1427 \\
Unit 3 & 135 & 135 & 150 & 150 & 136 & 148 & 150 & 150 & 150 & 52 & 150 & 1454 \\
Unit 4 & 135 & 133 & 150 & 145 & 120 & 150 & 150 & 150 & 150 & 71 & 150 & 1433 \\
\midrule
\textbf{Total} & \textbf{540} & \textbf{541} & \textbf{600} & \textbf{594} & \textbf{456} & \textbf{598} & \textbf{600} & \textbf{600} & \textbf{600} & \textbf{370} & \textbf{600} & \textbf{6099} \\
\bottomrule
\end{tabular}
\caption{\textbf{Statistics of the test set.} The test set comprises 10 years of authentic TCM examination papers, with each year covering 4 units that span the full spectrum of knowledge in traditional Chinese medicine. All data have been rigorously annotated and quality-controlled by a team of professional physicians.}
\label{tab:stastic_test}
\end{table*}
The National Qualification Examination for Traditional Chinese Medicine (TCM) Practitioners, recognized as the authoritative professional certification in China, integrates the full scope of TCM across 18 official textbooks, making it the field’s de facto gold standard. TCM-Eval is designed to rigorously evaluate the reasoning, memorization, and application abilities of models in the TCM domain by leveraging this gold standard. Unlike the TCM-Corpus, which emphasizes knowledge acquisition and exposure to diverse question formats, TCM-Eval exclusively prioritizes authenticity, reliability, and dynamic renewal to ensure fair and representative benchmarking.

\paragraph{Source of Examination Items}
To reflect real-world competence requirements, the test items were derived from ten years of authentic questions in the National Qualification Examination for TCM Practitioners. This examination is recognized as the gold standard for evaluating practitioners’ mastery of TCM knowledge and clinical reasoning skills. The coverage spans classical theories, diagnostic methodologies, herbal prescriptions, acupuncture techniques, and integrative applications with modern medicine, thereby offering a comprehensive basis for evaluating model performance.

\paragraph{Dynamic Refresh Mechanism}

A potential concern for long-term benchmarks in domain-specific large models is the risk of data leakage, which could compromise the validity of test results. To address this, we collaborated with a panel of licensed TCM professionals to periodically design and release novel examination sets that closely follow the style and difficulty of the official examination, yet are never exposed in public training corpora. This dynamic update mechanism ensures that the benchmark remains robust against memorization and provides a continually challenging environment for evaluating genuine reasoning ability.

\subsection{TCM-Corpus} \label{sec:train_data_collection}

The domain of Traditional Chinese Medicine (TCM) encompasses not only a broad spectrum of knowledge ( including both classical TCM theories and modern medical concepts ) but also constitutes a self-contained system of dialectical reasoning, which introduces unique challenges for model training. Drawing inspiration from the authentic learning process of human students, we design a data collection pipeline that systematically extracts knowledge-oriented question–answer pairs from textbooks and acquires mock examination items through automated procedures, thereby emulating the dual processes of study and practice in TCM education. Furthermore, we propose a self-iterative framework to generate high-quality Chain-of-Thought (CoT) data.

\paragraph{Textbook Extraction}
Due to the scarcity of high-quality datasets in the TCM domain and the unreliability of online sources, only the 18 authoritative textbooks were utilized to ensure data quality. High-quality text data were extracted using automated scanning tools and OCR. Since the raw OCR output contained numerous irrelevant and noisy characters, manually designed rules were applied to remove extraneous elements such as headers, footers, and other artifacts.
\begin{table*}[!htbp]
\small
\centering
\begin{tabular}{@{}llccc@{}} 
\toprule
\textbf{Dataset} & \textbf{Description} & \textbf{Source} & \textbf{Items} & \textbf{Tokens} \\ 
\midrule
TCM Internal Medicine & Diagnosis and treatment of internal diseases & \imgintext{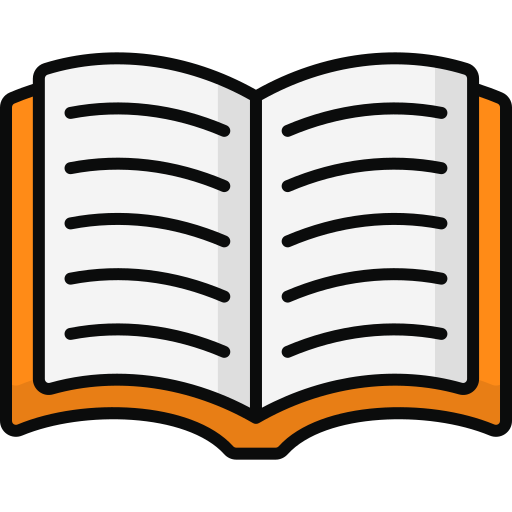}~\texttt{+}~\imgintext{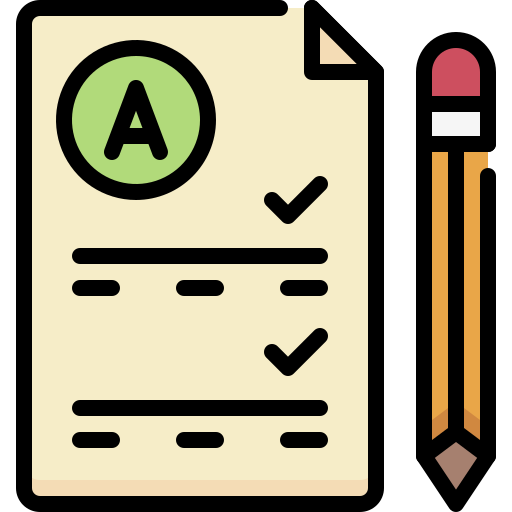} & 46,599 & 20.6M \\
TCM Surgery & Diagnosis and treatment of surgical diseases & \imgintext{fig/textbook.png}~\texttt{+}~\imgintext{fig/exam.png} & 45,031 & 18.2M \\
Infectious Diseases & Prevention and treatment of infectious diseases & \imgintext{fig/textbook.png}~\texttt{+}~\imgintext{fig/exam.png} & 39,953 & 15.4M \\
TCM Pediatrics & Diagnosis and treatment of pediatric diseases & \imgintext{fig/textbook.png}~\texttt{+}~\imgintext{fig/exam.png} & 35,609 & 14.6M \\
Chinese Materia Medica & Knowledge of Chinese medicinal herbs & \imgintext{fig/textbook.png}~\texttt{+}~\imgintext{fig/exam.png} & 31,620 & 13.1M \\
Health Law and Regulations & Medical laws and regulations & \imgintext{fig/textbook.png}~\texttt{+}~\imgintext{fig/exam.png} & 29,110 & 11.1M \\
TCM Diagnostics & Diagnostic methods in TCM & \imgintext{fig/textbook.png}~\texttt{+}~\imgintext{fig/exam.png} & 28,389 & 12.0M \\
Basic Theory of TCM & Fundamentals of Traditional Chinese Medicine & \imgintext{fig/textbook.png}~\texttt{+}~\imgintext{fig/exam.png} & 24,690 & 9.6M \\
Acupuncture and Moxibustion & Acupuncture, moxibustion, and Tuina & \imgintext{fig/textbook.png}~\texttt{+}~\imgintext{fig/exam.png} & 22,974 & 9.7M \\
Chinese Herbal Formulas & Formula composition and compatibility & \imgintext{fig/textbook.png}~\texttt{+}~\imgintext{fig/exam.png} & 15,051 & 6.6M \\
TCM Ethics & Medical ethics in TCM & \imgintext{fig/textbook.png}~\texttt{+}~\imgintext{fig/exam.png} & 14,688 & 5.6M \\
TCM Gynecology & Gynecological diseases & \imgintext{fig/textbook.png}~\texttt{+}~\imgintext{fig/exam.png} & 11,952 & 5.6M \\
Warm-Febrile Diseases & Theory and treatment of warm-febrile diseases & \imgintext{fig/textbook.png} & 10,106 & 3.7M \\
Shang Han Lun & Theory of cold damage diseases & \imgintext{fig/textbook.png} & 7,823 & 2.8M \\
Jin Gui Yao Lue & Theory of miscellaneous diseases & \imgintext{fig/textbook.png} & 6,465 & 2.4M \\
Huangdi Neijing & The Yellow Emperor's Inner Canon & \imgintext{fig/textbook.png} & 5,688 & 2.2M \\
Other & Other subjects & \imgintext{fig/exam.png} & 9,059 & 5.8M \\ 
\midrule
\textbf{Total} & \multicolumn{2}{c}{} & \textbf{384,807} & \textbf{159M} \\ 
\bottomrule
\end{tabular}
\caption{\textbf{Statistics of the training set.} This dataset provides comprehensive knowledge coverage across 16 Traditional Chinese Medicine (TCM) domains, constructed from 18 authoritative textbooks and 1601 manually collected mock exams. In the "Source" column, the \imgintext{fig/textbook.png} icon indicates data derived from textbooks, while the \imgintext{fig/exam.png} icon represents data from mock exams.}
\label{tab:training_dataset_stats}
\end{table*}

Using the collected textbook data, we segmented the textbook data into coherent blocks. to achieve comprehensive coverage of knowledge points. The segmentation was guided by chapter titles and controlled by text length to maintain readability. Subsequently, manual inspection was performed to refine the boundaries, ensuring that individual knowledge points remained intact and were not fragmented across multiple segments.

To enable the model to acquire TCM knowledge progressively, we designed prompts to guide DeepSeek-v3 in generating question–answer pairs of varying difficulty from textbook segments. Since TCM knowledge includes a substantial proportion of memorization-based content, fill-in-the-blank questions were used as the primary format. To enhance diversity and align with examination styles, we additionally generated multiple-choice questions.

\paragraph{Mock Exam Harvesting}
To familiarize the model with the format and style of China’s TCM Practitioner Qualification Examination, automated web crawlers were employed to collect over 60,000 candidate mock exam questions from university-hosted forums and platforms. The raw data contained invalid options, invalid answers, and duplicates. Rule-based filtering was applied to remove incomplete or malformed items, while deduplication based on question-stem similarity was used to reduce redundancy.

\paragraph{SI-CoTE}
To training our reasoning model — \textbf{ZhiMingTang}, we proposed \textbf{Self-Iterative Chain-of-Thought Enhancement (SI-CoTE)} framework. This framework is built upon the Deepseek-R1  \cite{guo2025deepseek} base model, denoted as $\mathcal{M}_0$. At its core, the SI-CoTE methodology employs a phased, iterative process to efficiently convert the original Question-Answer (QA) pair dataset, $\mathcal{D}_{\text{QA}}$, into an enhanced dataset containing high-quality CoT data, thereby driving the evolution of the model's capabilities.

Initially, we partition the original QA dataset into $K$ disjoint subsets:

\begin{equation}
    \mathcal{D}_{\text{QA}} = \mathcal{D}_1 \cup \mathcal{D}_2 \cup \dots \cup \mathcal{D}_K
    \label{eq:partition}
\end{equation}

Our training process then unfolds iteratively over these subsets. In the $k$-th iteration (for $k=1, \dots, K$), we focus on generating high-quality CoT for the QA pairs $(Q_i, A_i^*) \in \mathcal{D}_k$.

Specifically, in the $k$-th iteration, we leverage the model from the previous iteration, $\mathcal{M}_{k-1}$, where $\mathcal{M}_0$ is the initial base model, to generate candidate CoT and answer pairs $(T_i, A_i)$ for each question $Q_i$ in subset $\mathcal{D}_k$. The generated answers are then filtered using a verification function. This \textbf{Rejection Sampling} step retains a candidate if and only if its generated answer $A_i$ is consistent with the ground-truth answer $A_i^*$. This condition is formally expressed as:

\begin{equation}
    V(A_i, A_i^*) = 1
    \label{eq:verification}
\end{equation}

For questions where the model fails to produce a correct answer after multiple attempts, these "hard cases" are identified and annotated with high-quality CoT by medical experts. Through this combined approach of machine generation-verification and human-in-the-loop assistance, we construct a high-quality CoT dataset, $\mathcal{D}_{\text{CoT}, k}$, for the corresponding data subset $\mathcal{D}_k$.

The key to this framework lies in the accumulation of data and the iterative evolution of the model. The newly generated CoT data from each iteration is aggregated into a cumulative SFT training set. At the conclusion of the $k$-th iteration, this set is updated as follows:

\begin{equation}
    \mathcal{D}_{\text{SFT}}^{(k)} = \bigcup_{j=1}^k \mathcal{D}_{\text{CoT}, j}
    \label{eq:update_sft}
\end{equation}

We then fine-tune the base model $\mathcal{M}_0$ on this continuously expanding and refined dataset to yield a more capable next-generation model, $\mathcal{M}_k$:

\begin{equation}
    \mathcal{M}_k = \text{SFT}(\mathcal{M}_0, \mathcal{D}_{\text{SFT}}^{(k)})
    \label{eq:finetune}
\end{equation}

Having learned the reasoning patterns from the first $k$ subsets, the enhanced model $\mathcal{M}_k$ exhibits stronger reasoning capabilities and a higher success rate when processing the subsequent subset, $\mathcal{D}_{k+1}$. This process continues until all $K$ subsets have been processed. The final model, \textbf{ZhiMingTang}, denoted as $\mathcal{M}_K$, is trained on the complete, high-quality CoT dataset $\mathcal{D}_{\text{SFT}}^{(K)}$. Through this batch-wise, iterative enhancement approach, the SI-CoTE framework achieves a synergistic evolution of model capability and data quality.

\subsection{Quality Control} \label{sec:data_quality_control}

To ensure correctness of TCM-Eval and to improve the coherence of TCM-Corpus, we implement a multi-stage quality control protocol, supporting reliable model evaluation and training.

For TCM-Eval, each test item was manually collected and curated by a dedicated team of medical experts. Every question underwent multi-round annotation and verification by independent reviewers to guarantee accuracy in both stems and answers. Items with ambiguous wording or multiple plausible answers were excluded to avoid confounding evaluation outcomes. This strict quality assurance process ensures that the final test dataset is composed of high-quality, unambiguous, and exam-standard questions, suitable for serving as a reliable yardstick of model performance.

For TCM-Corpus, a model-in-the-loop validation procedure was introduced \cite{liu-etal-2020-towards-conversational, cheng-etal-2025-toolspectrum, liu2025repodebugrepositorylevelmultitaskmultilanguage}. Items consistently answered correctly by the model were retained as high-confidence samples, while items answered incorrectly were flagged for stricter review, including both automated heuristics and human inspection. This hybrid validation process yielded a final curated dataset.

\subsection{Data Statistics} \label{sec:data_stat}

We conducted a detailed statistical analysis of the training and test sets for \texttt{\TCMBENCH}, as shown in Table \ref{tab:training_dataset_stats} and Table \ref{tab:stastic_test}. The training set covers 16 knowledge domains in traditional Chinese medicine, constructed from 1,601 sets of practice questions and 18 authoritative official textbooks, comprising a total of 384,807 instruction-tuning samples and 159 million tokens. These data enable the model to acquire foundational knowledge and key examination topics in TCM during the post-training phase.

The test set consists of 6,099 questions, spanning 10 years of official exam questions, uniformly distributed across four modules to assess different model capabilities. To prevent data leakage and ensure the integrity of model evaluation, a team of professional TCM practitioners manually curated 600 high-quality questions. This dynamic test set will be continuously maintained and updated.

\begin{table*}[!ht]
\centering
\small 
\renewcommand\tabcolsep{5pt} 
\renewcommand\arraystretch{1.4} 

\resizebox{\linewidth}{!}{%
\begin{tabular}{lccccccccccccc}
\toprule
\rowcolor{gray!25}
\textbf{Model} & \textbf{Size} & \textbf{2003} & \textbf{2004} & \textbf{2007} & \textbf{2008} & \textbf{2009} & \textbf{2012} & \textbf{2013} & \textbf{2016} & \textbf{2022} & \textbf{2024} & \textbf{HC} & \textbf{Overall} \\
\midrule
\rowcolor{gray!15}
\multicolumn{14}{c}{\textit{\textbf{Open-Sourced}}} \\
\midrule
Qwen3-8B & 8B  & 69.46 & 67.16 & 64.36 & 61.47 & 71.05 & 63.82 & 64.66 & 65.77 & 63.38 & 76.76 & 68.17 & 66.45\\
Qwen3-14B & 14B & 79.33 & 80.60 & 78.89 & 76.88 & 77.19 & 76.88 & 78.89 & 81.98 & 77.42 & 80.00 & 75.33 & 78.35 \\
Qwen3-32B & 32B & 85.29 & 86.01 & 84.97 & 86.13 & 85.09 & 85.26 & 83.42 & 88.51 & 86.29 & 83.51 & 81.50 & 85.04 \\
Qwen3-30B-A3B & 30B & 57.36 & 61.75 & 60.47 & 65.24 & 62.28 & 62.98 & 59.90 & 68.92 & 55.69 & 68.11 & 53.83 & 61.07 \\
Qwen3-235B-A22B & 235B & \second{95.34} & 95.52 & 92.40 & \second{95.55} & \second{96.05} & 92.63 & \third{94.64} & 92.57 & \second{93.81} & 87.84 & \second{91.00} & \second{93.52} \\
GPT-oss-20B & 20B & 35.57 & 37.69 & 36.82 & 39.90 & 36.40 & 37.19 & 37.19 & 40.32 & 41.47 & 45.14 & 40.17 & 38.72 \\
GPT-oss-120B & 120B & 54.38 & 58.40 & 52.87 & 54.28 & 55.70 & 54.44 & 57.45 & 59.23 & 60.20 & 59.19 & 56.56 & 56.49 \\
DS-Qwen-7B & 7B & 27.75 & 33.21 & 28.21 & 30.14 & 28.29 & 30.49 & 32.83 & 32.43 & 33.11 & 33.78 & 33.83 & 31.25 \\
DS-Qwen-14B & 14B & 74.30 & 77.24 & 77.20 & 75.00 & 75.88 & 74.54 & 73.53 & 78.38 & 77.42 & 72.43 & 75.33 & 75.60 \\
DS-Qwen-32B & 32B & 77.09 & 82.28 & 78.55 & 79.11 & 80.92 & 75.38 & 79.40 & 83.78 & 80.94 & 74.05 & 79.17 & 79.17 \\
DS-Llama-8B & 8B & 28.12 & 27.61 & 25.34 & 27.74 & 27.19 & 30.49 & 32.33 & 32.33 & 27.70 & 30.27 & 30.17 & 28.74 \\
DS-Llama-70B & 70B & 56.42 & 58.96 & 56.42 & 58.90 & 55.92 & 55.28 & 58.29 & 56.98 & 61.04 & 55.41 & 58.00 & 57.54 \\
Deepseek-V3.1 & 685B & 91.43 & 93.66 & 91.05 & 91.78 & 91.67 & 90.79 & 90.62 & 90.54 & 88.96 & 88.65 & 86.83 & 90.54 \\
Deepseek-R1 & 671B & 80.91 & 93.28 & 90.88 & 90.41 & 88.82 & 89.45 & 89.78 & 90.32 & 89.30 & 85.68 & \third{90.56} & 89.04 \\
Llama-3.1-8B & 8B & 47.11 & 51.49 & 44.76 & 48.46 & 47.59 & 48.58 & 48.58 & 48.87 & 50.50 & 48.11 & 50.83 & 48.66 \\
Llama-3.3-70B & 70B & 65.74 & 71.08 & 69.43 & 73.12 & 70.39 & 65.66 & 65.16 & 68.24 & 69.23 & 67.03 & 67.17 & 68.38 \\
Ministral-8B & 8B & 33.40 & 34.96 & 33.85 & 34.49 & 33.19 & 35.81 & 34.68 & 31.53 & 36.35 & 28.92 & 35.68 & 33.90 \\
MedGemma-4B & 4B & 30.04 & 34.33 & 29.47 & 31.56 & 32.16 & 30.81 & 31.64 & 32.35 & 32.09 & 35.89 & 34.39 & 32.25 \\
MedGemma-27B & 27B & 44.67 & 51.96 & 43.05 & 48.37 & 49.45 & 44.46 & 48.24 & 48.75 & 50.08 & 57.53 & 50.17 & 48.79 \\
Baichuan-M2 & 32B & 51.58 & 55.04 & 60.30 & 63.18 & 56.36 & 60.13 & 54.61 & 63.06 & 50.00 & 54.32 & 44.67 & 55.63 \\
ShiZhenGPT-7B & 7B & 79.58 & 83.69 & 82.69 & 82.77 & 81.43 & 81.50 & 83.74 & 81.92 & 82.53 & 75.64 & 82.64 & 81.65 \\
ShiZhenGPT-32B & 32B & 89.33 & 93.22 & 90.14 & 93.85 & 90.62 & 88.89 & 90.02 & 91.16 & 89.57 & 87.29 & 87.78 & 90.17 \\
\midrule
\rowcolor{gray!15}
\multicolumn{14}{c}{\textit{\textbf{API-Based}}} \\
\midrule
GLM-4.5 & 358B & 85.29 & 86.19 & 84.12 & 86.64 & 85.53 & 86.93 & 85.76 & 89.64 & 82.27 & 85.14 & 79.33 & 85.03 \\
GLM-4.5-Air & 110B & 75.98 & 79.66 & 77.87 & 79.62 & 79.61 & 78.39 & 79.40 & 82.66 & 75.42 & 77.84 & 73.00 & 77.99 \\
MiniMax-M1 & 456B & 71.32 & 76.12 & 74.66 & 76.88 & 75.88 & 75.04 & 73.70 & 76.58 & 71.91 & 74.32 & 73.32 & 74.52 \\
GPT-4o & -- & 71.14 & 77.24 & 71.62 & 73.46 & 75.00 & 73.20 & 77.89 & 77.48 & 76.25 & 67.30 & 75.32 & 74.17 \\
GPT-4.1 & -- & 73.18 & 78.17 & 71.62 & 72.09 & 73.46 & 72.70 & 75.21 & 78.83 & 75.92 & 72.43 & 72.83 & 74.22 \\
Baichuan4 & -- & 92.92 & 93.10 & \third{92.91} & 93.32 & 92.54 & 92.29 & 91.44 & 91.67 & 91.14 & 83.78 & 88.33 & 91.22 \\
LongCat-Flash-Chat & 560B & 91.25 & \third{95.71} & 92.23 & 92.98 & 91.89 & 91.96 & 91.46 & 92.12 & 91.25 & 88.89 & 87.83 & 91.60 \\
Hunyuan-T1 & -- & 91.25 & 92.91 & 92.74 & \third{95.38} & \third{93.86} & \third{92.96} & 92.80 & \third{94.37} & \third{92.14} & \third{91.08} & 88.00 & 92.47 \\
Kimi-K2-Instruct & 1T & \third{94.79} & \best{97.76} & \second{95.10} & \third{95.38} & \second{96.05} & \second{94.47} & \second{94.97} & \second{95.27} & \third{92.14} & \second{92.16} & 79.70 & \third{93.30} \\
Spark-4.0-Ultra & -- & 88.83 & 93.10 & 91.05 & 90.58 & 89.91 & 89.78 & 91.29 & 91.44 & 90.47 & 85.41 & 87.17 & 90.02 \\
Ernie-x1-turbo-32k & -- & 87.90 & 90.11 & 86.82 & 85.96 & 91.45 & 85.43 & 84.59 & 84.68 & 83.11 & 90.54 & 89.43 & 87.27 \\
\midrule
\textbf{ZMT-M1 (Ours)} & 671B & \best{97.02} & \second{96.27} & \best{96.96} & \best{96.58} & \best{96.93} & \best{95.98} & \best{95.98} & \best{95.95} & \best{94.48} & \best{97.03} & \best{95.67} & \best{96.26} \\
\bottomrule
\end{tabular}%
}
\caption{
    \textbf{Comparison of open-sourced and API-based models across years}, with an \textbf{HC} representing the Hand-Crafted set and \textbf{Overall} column reporting the average score. The \colorbox{bestcolor}{best}, \colorbox{secondcolor}{second-best}, and \colorbox{thirdcolor}{third-best} results in each column are marked with purple, orange, and gray backgrounds, respectively.
}
\label{tab:main_results}
\end{table*}

\section{Experiment}
\subsection{Setup}
\paragraph{Models}

Following previous work \cite{chen2025shizhengptmultimodalllmstraditional, cheng-etal-2025-toolspectrum}, we conducted extensive experiments on a wide range of \textbf{open-source} and \textbf{API-based} models. 

Specifically, the \textbf{open-sourced} models encompass various series with different parameter scales. For general-purpose large language models, our selection includes Qwen3 series (8B, 14B, 32B) \cite{yang2025qwen3}, Llama-3 series (8B, 70B) \cite{grattafiori2024llama}, Deepseek series (V3.1, R1) \cite{liu2024deepseek, guo2025deepseek}, Baichuan-M2 (32B) \cite{dou2025baichuan}, and Mistral-8B\footnote{\url{https://mistral.ai/news/ministraux}}, GPT-oss series (20B, 120B) \cite{openai2025gptoss120bgptoss20bmodel}, the DS-Qwen series (7B, 14B, 32B), and the DS-Llama series (8B, 70B). To account for domain-specific applications, we also incorporated models known for their performance in the medical field, namely the MedGemma series (4B, 27B) \cite{sellergren2025medgemmatechnicalreport} and the ShiZhenGPT series (7B, 32B) \cite{chen2025shizhengptmultimodalllmstraditional}.

For the \textbf{API-based} models, our evaluation includes services from various providers. These include GLM-4 series (GLM-4.5, GLM-4.5-Air) \cite{5team2025glm45agenticreasoningcoding}, Kimi-K2-Instruct \cite{kimiteam2025kimik2openagentic}, Baichuan4\footnote{\url{https://platform.baichuan-ai.com/}}, Ernie-x1-turbo-32k \cite{}, Spark-4.0-Ultra\footnote{\url{https://xinghuo.xfyun.cn/}}, MiniMax-M1 \cite{minimax2025minimaxm1scalingtesttimecompute}, LongCat-Flash-Chat \cite{meituan2025longcatflashtechnicalreport}, and GPT-4 series\footnote{\url{https://chatgpt.com/}} (GPT-4o, GPT-4.1).

\paragraph{Benchmarks}

As introduced in Section \ref{sec:test_data_collection}, we constructed \texttt{\TCMBENCH} to comprehensively evaluate the knowledge and reasoning capabilities of LLMs in the field of TCM. We use this as our primary benchmark, testing the accuracy of each model on different question sets.

\paragraph{Implemental Details}

To ensure the reproducibility and fairness of our results, we standardized the inference and fine-tuning procedures for all models. For inference, we employed the efficient vLLM framework \cite{kwon2023efficientmemorymanagementlarge} as the unified engine for all open-sourced models. For non-reasoning or deterministic generation tasks, we set the decoding temperature to 0 with both top-p and top-k configured to 0.1, whereas for tasks requiring reasoning, the temperature was set to 0.6. In the Supervised Fine-Tuning (SFT) phase, we customized the SFTTrainer from the Hugging Face Transformers library \cite{wolf-etal-2020-transformers} and adopted the parameter-efficient Low-Rank Adaptation (LoRA) \cite{hu2022lora}. All experiments were conducted on a single node equipped with eight H20 GPUs, each with 141 GB of VRAM, and the entire training phase consumed approximately 7,000 GPU-hours.

\subsection{Main Results}


As shown in Table \ref{tab:main_results}, we conducted experiments on a wide range of models using the past ten years of official examination questions as well as Human-Crafted items, and subsequently computed the Overall score. From these results, we can draw the following conclusions:

\paragraph{\textit{ZMT-M1 demonstrated the most outstanding performance in the qualification examination for TCM practitioners.}} As shown in Table \ref{tab:main_results}, it achieved the highest scores in 10 out of 11 test subsets. With the inclusion of updated test items (such as the 2024 official exam questions and Human-Crafted items), ZMT-M1’s advantage has further expanded. Because these items carry a lower risk of data leakage, they provide a more authentic reflection of the model’s true capabilities, thereby underscoring ZMT-M1’s robustness and authority in the field of TCM.


\paragraph{\textit{Compared with Deepseek-R1, our improvements are substantial, fully demonstrating the high quality of the training corpus.}} Although the untrained Deepseek-R1 already achieved an Overall score of 89.04, it still lagged behind the state-of-the-art models. After training, however, ZMT-M1’s score increased to 96.17, representing a gain of 7.13 points. This indicates that our training corpus—comprising textbooks and mock exam questions—provides comprehensive coverage of knowledge points in Traditional Chinese Medicine and delivers targeted enhancement of the model’s examination performance.


\paragraph{\textit{Models developed by Chinese enterprises or academic institutions tend to perform better in the field of TCM.}} For instance, as shown in the table, although GPT-4o and GPT-4.1 are among the strongest models across most other domains, their performance in TCM is even inferior to that of Qwen3-8B. We attribute this primarily to differences in the proportion of TCM-related content within the training corpora, as well as variations in the models’ capabilities for processing Chinese.

\section{Analysis}
In this section, we conduct a comprehensive analysis to answer the three research questions \textbf{RQ1}: \textit{How does the model perform across different sub-domains or tasks within traditional Chinese medicine?} (Sec \ref{sec:rq1}) \textbf{RQ2}: \textit{Is the SI-CoTE training approach universally effective across different model architectures?} (Sec \ref{sec:rq2}) \textbf{RQ3}: \textit{Through case studies, does the model genuinely acquire knowledge specific to traditional Chinese medicine after training?} (Sec \ref{sec:rq3})

\subsection{Analysis of Model Performance Across Various Aspects of TCM} \label{sec:rq1}

\begin{figure}[!htbp]
    \centering
    \includegraphics[width=\linewidth]{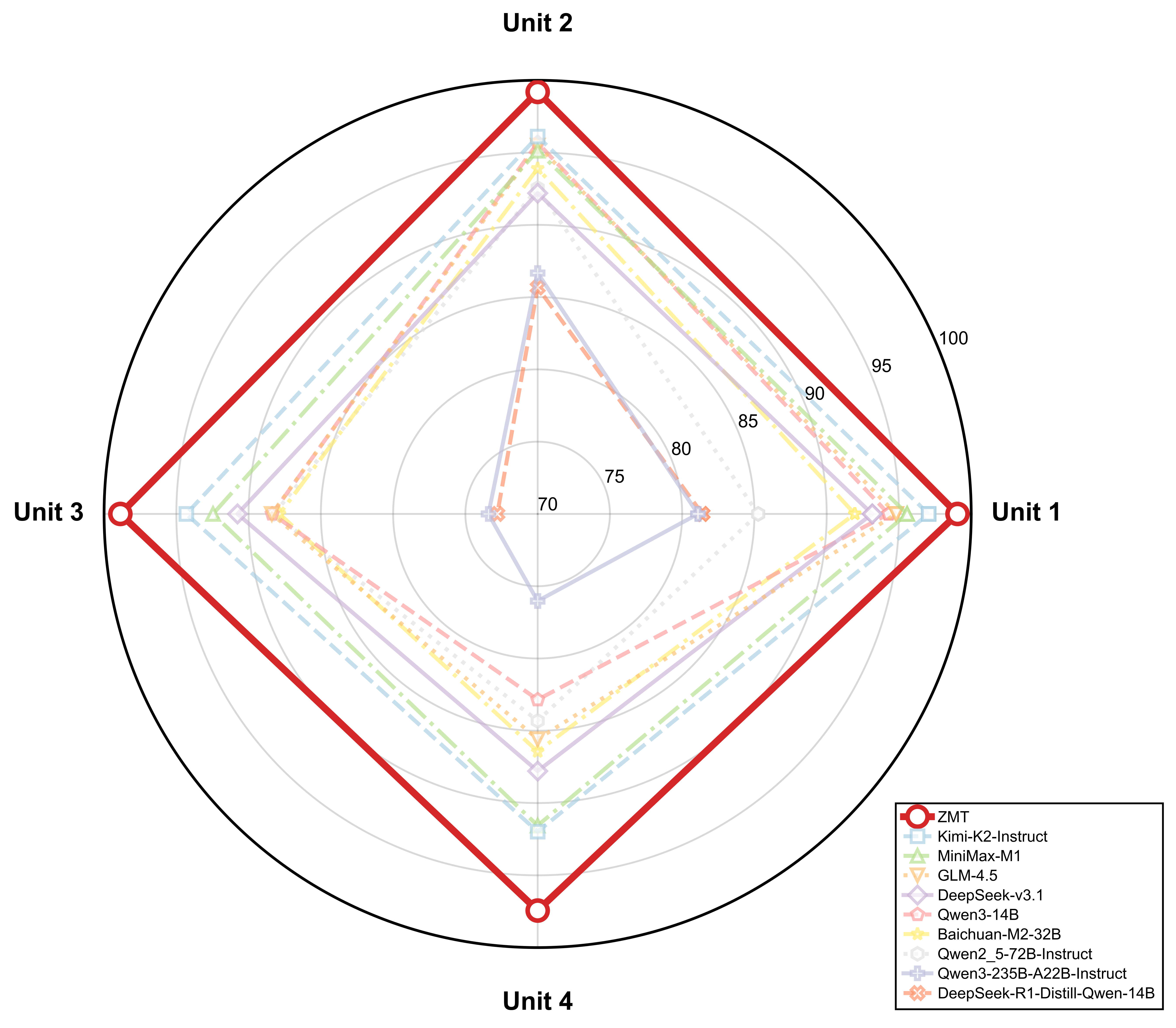}
    \caption{\textbf{Performance distribution of ZMT-M1 and baseline models across four evaluation units.} ZMT-M1 exhibits the most balanced performance profile with stable scores on all units. In contrast, baseline models demonstrate significant performance variations, particularly weaker on Units 3 and 4 compared to Unit 2.}
    \label{fig:rq1}
\end{figure}

As shown in Table~\ref{tab:stastic_test}, \texttt{\TCMBENCH} is divided into four units, each designed to assess a distinct capability within the domain of TCM. We have separately evaluated and compared the performance of various models across these four units. Figure~\ref{fig:rq1} reveals a significant variance in the models' abilities across different dimensions of TCM. Specifically, the scores in Unit 4 are substantially lower than those in Unit 2, indicating that existing models have a notable deficiency in clinical application skills. In contrast, our proposed ZMT-M1 model not only surpasses other models in every unit but also demonstrates the most balanced proficiency, achieving the lowest score variance. This superior and well-rounded performance is attributed to the comprehensive coverage of knowledge points during our model's training phase.

\subsection{Generalizability of the SI-CoTE} \label{sec:rq2}
\begin{table}[!ht]
    \centering
    \renewcommand{\arraystretch}{1.3}
    \setlength{\tabcolsep}{3mm}
    \resizebox{\linewidth}{!}{%
    \begin{tabular}{@{}l|c|*{2}{S[table-format=2.2]}|S[table-format=3.2]@{}}
    \toprule
    \textbf{Model} & \textbf{Size} & {\textbf{OA (original)}} & {\textbf{OA (w/ fine-tuning)}} & {\textbf{$\Delta_{rel}^\%$}} \\
    \midrule
    Qwen3-8B & 8B & 66.45 & 82.91 & {\cellcolor[RGB]{155,217,155}$\uparrow 24.77$} \\
    Qwen3-14B & 14B & 78.35 & 86.79 & {\cellcolor[RGB]{205,244,205}$\uparrow 10.77$} \\
    Qwen3-32B & 32B & 85.04 & 88.21 & {\cellcolor[RGB]{245,255,245}$\uparrow 3.73$} \\
    \midrule
    DS-Qwen-7B & 7B & 31.25 & 81.52 & {\cellcolor[RGB]{132,203,132}$\uparrow 160.86$} \\
    DS-Llama-8B & 8B & 28.74 & 77.35 & {\cellcolor[RGB]{110,190,110}$\uparrow 169.14$} \\
    DS-Qwen-14B & 14B & 75.60 & 86.31 & {\cellcolor[RGB]{180,231,180}$\uparrow 14.17$} \\
    DS-Qwen-32B & 32B & 79.17 & 87.18 & {\cellcolor[RGB]{225,250,225}$\uparrow 10.12$} \\
    \bottomrule
    \end{tabular}%
    }
    \caption{
        \textbf{Comparison of overall accuracy before and after fine-tuning models on Chain-of-Thought training data generated with SI-CoTE. }
    }
    \label{tab:RQ2}
    \end{table}

\begin{figure*}[!htbp]
    \centering
    \includegraphics[width=\textwidth]{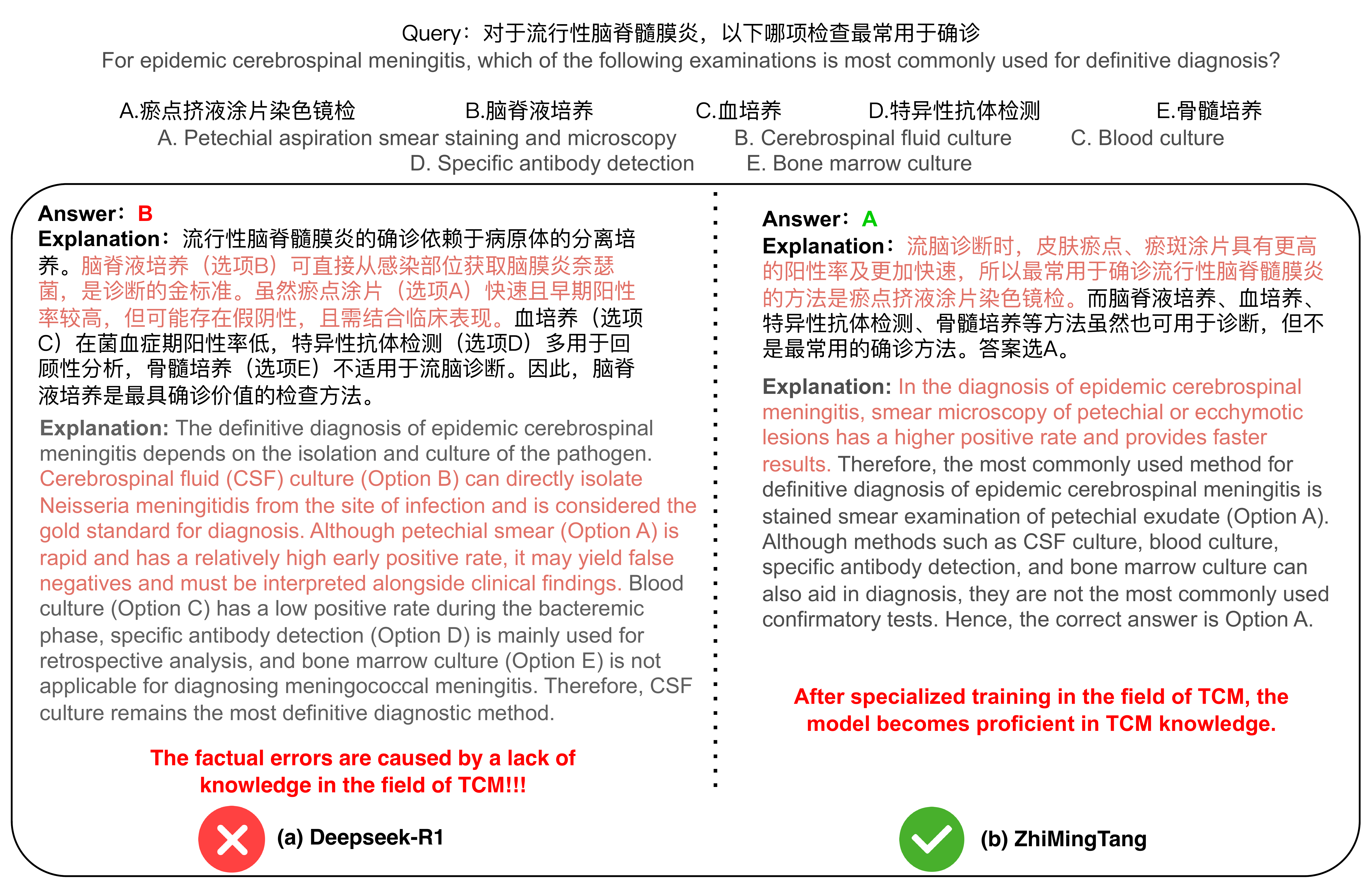}
    \caption{\textbf{Case Study.} This study demonstrates that after specialized training in TCM using the SI-CoTE method, the ZhiMingTang exhibits a more comprehensive knowledge base than the general-purpose Deepseek-R1. Critically, it effectively integrates clinical case experience to provide more precise and professionally accurate responses.}
    \label{fig:RQ3}
\end{figure*}
To ascertain the efficacy of the CoT augmented training data from Section~\ref{sec:method}, we conducted fine-tuning experiments on open-source models with known deficiencies in the TCM domain. As presented in Table~\ref{tab:RQ2}, fine-tuning the Qwen3 and Deepseek-Distill series with our data yielded significant performance enhancements. The improvements were particularly pronounced for smaller models, which saw performance boosts of up to 135\%, rivaling the capabilities of API-based models. These findings underscore the high quality and broad coverage of the data produced by our SI-CoTE iterative method, confirming its effectiveness in advancing model performance in the TCM field.


\subsection{Case Study} \label{sec:rq3}

This section presents a qualitative, side-by-side comparison of DeepSeek-R1 and ZMT-M1 (Figure~\ref{fig:RQ3}), highlighting key differences in their approach to and understanding of TCM. The question shown in the figure combines textbook knowledge with clinical diagnostic experience. It is evident that ZMT-M1 has a broader knowledge base within the scope of TCM compared to DeepSeek-R1. ZMT-M1 demonstrates a better grasp of textbook knowledge and integrates real clinical experience. Additionally, ZMT-M1 provides more concise and accurate answers, reflecting its greater expertise in the field of TCM.

\section{Conclusion}
This work addresses critical gaps in LLM applications for TCM. We introduce \TCMBENCH, the first dynamic, expert-validated benchmark sourced from national licensing examinations, establishing a rigorous 'gold standard' for robust LLM evaluation. To foster advanced reasoning, our novel Self-Iterative Chain-of-Thought Enhancement (SI-CoTE) framework enables autonomous CoT generation and validation, driving a virtuous cycle of data and model co-evolution. Leveraging this, our SOTA LLM, ZMT-M1, achieves an unprecedented 96.32\% on \TCMBENCH, far surpassing human practitioner pass rates and setting a new benchmark for expert-level knowledge and reasoning. By open-sourcing these resources and maintaining a dynamic public leaderboard, we aim to catalyze future research and accelerate TCM's AI-driven modernization.

\bibliography{custom}



\end{document}